\documentclass[11pt,a4paper]{article}
\usepackage[hyperref]{nlp4MusA}
\usepackage{times}
\usepackage{latexsym}
\usepackage[pdftex]{graphicx}
\usepackage{url}
\usepackage{multirow}
\usepackage{graphicx}
\usepackage{booktabs}
\usepackage{kotex}
\usepackage{lipsum}  

\usepackage{amssymb}
\usepackage{amsfonts} 

\aclfinalcopy 

\title{Music Playlist Title Generation: A Machine-Translation Approach}

\author{SeungHeon Doh$^{1}$\hspace{1cm} Junwon Lee$^{2}$ \hspace{1cm} Juhan Nam$^{1}$ \\
  Graduate School of Culture Technology, KAIST, South Korea$^{1}$ \\
  Department of Electrical Engineering, KAIST, South Korea$^{2}$ \\
  \texttt{\{seungheondoh,james39,juhan.nam\}@kaist.ac.kr} \\
  }
\date{}
\begin{document}
\maketitle
\begin{abstract}

We propose a machine-translation approach to automatically generate a playlist title from a set of music tracks. We take a sequence of track IDs as input and a sequence of words in a playlist title as output, adapting the sequence-to-sequence framework based on Recurrent Neural Network (RNN) and Transformer to the music data. Considering the orderless nature of music tracks in a playlist, we propose two techniques that remove the order of the input sequence. One is data augmentation by shuffling and the other is deleting the positional encoding. We also reorganize the existing music playlist datasets to generate phrase-level playlist titles. The result shows that the Transformer models generally outperform the RNN model. Also, removing the order of input sequence improves the performance further.


\end{abstract}

\section{Introduction}
Music playlists have gained progressively more importance in music streaming services. A playlist represents a group of music tracks that shares similar genre, mood or musical context. When a new playlists is created by curators or users, or generated by recommender systems, they deliver messages about musical needs by providing playlist titles in a phrase \cite{pichl2015towards, dias2017manual}. However, it is not trivial to blend semantics of the music tracks and express them with a phrase. As a result, we often find noisy playlist titles which do not accord with the music tracks. 


\begin{figure}[t]
\centering
\includegraphics[width=\linewidth]{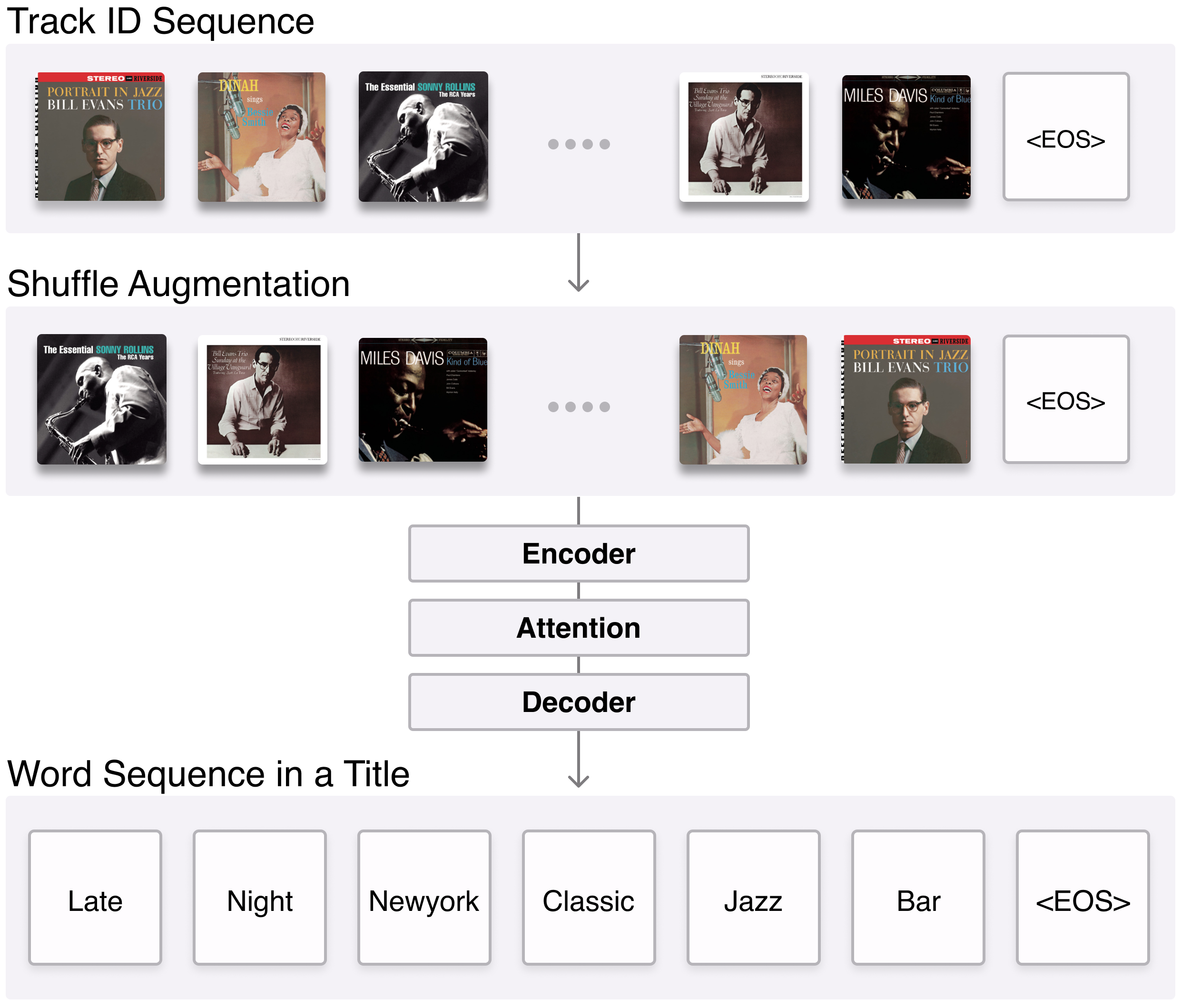}
\caption{Diagram of track ID sequence to word sequence in a title.}
\label{fig_1}
\end{figure}

A fundamental issue in automatic playlist title generation is to extract the common semantic features from the music tracks in a playlist, independent of the number of tracks. This issue has been addressed by representing a playlist with track embedding averaging \cite{hao2020using} or a sequential model \cite{choi2020prediction}. In \cite{hao2020using}, they treated playlists as the equivalent of phrases, and tracks as the equivalent of words. They then used the the word2vec model to learn the track embedding. In \cite{choi2020prediction}, they represented playlists and tracks as a matrix where the columns correspond to playlist IDs and the rows to track IDs. They then used a matrix factorization technique to learn the track embedding and, furthermore, applied an average or sequence model to predict high-level categorical labels. 

Another issue is to generate a natural word sequence (e.g., a phrase or a sentence) as a playlist title from the common semantics of music tracks. This sequence-to-sequence setting is similar to the machine translation task. 
Therefore it is natural to attempt the methods in machine translation, in particular, the encoder-decoder models \cite{bahdanau2014neural, vaswani2017attention}. 
This approach was previously attempted for playlist title generation \cite{sofia}. However, the model output was mostly tag-level titles (e.g., a single word or short phrase) rather than phrase-level titles, presumably because they used an unfiltered noisy dataset and a simple RNN model. Also, they used the track name as an input sequence. This input setting can confine tracks with similar names to have similar semantics, and also can learn the order of input sequence, which may be discarded in music playlists \cite{hao2020using}.    



\begin{figure*}[!t]
\centering
\includegraphics[width=\textwidth]{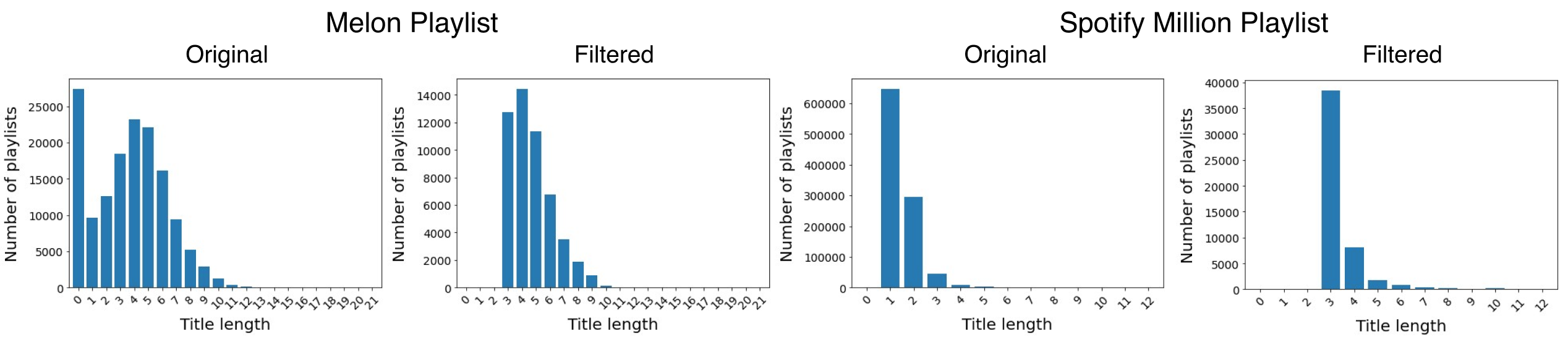}
\caption{Compare distribution of datasets(original, filtered), title length 0 means missing data, and title length 1 means tag-level title.}
\label{fig_2}
\end{figure*}

In this paper, we present another machine translation approach based on the encoder-decoder framework for automatic playlist title generation as illustrated in Figure \ref{fig_1}. Our contribution is as follows: (i) we compare two encoder-decoder models based on RNN and Transformer, (ii) we propose two simple techniques to make track ID sequence orderless and show that they improve the performance, and (iii) we propose a new data split by filtering existing playlist datasets and extracting phrase-level playlist title. 



\section{Dataset and Preprocessing}

We apply our proposed approach to two different datasets respectively: Melon Playlist Dataset (Melon) \cite{ferraro2021melon} and Spotify Million Playlist Dataset (MPD) \cite{chen2018recsys}. As our task is generating a playlist title in phrase for a given track ID sequence, we need a dataset of playlists that contains a pair of track ID sequence and title. Both Melon and MPD satisfy this requirement and support different languages (Korean and English). In Melon, playlist titles are written in both Korean and English (some of titles are mixed with both languages). In case of English words, normalization was done by substituting all characters with lowercase. Both of the languages were simply tokenized by white spaces.

In our task, an ideal playlist title is a phrase that incorporates common features among the songs in a playlist. However, Melon and MPD were originally constructed for automatic playlist continuation (APC) task and so they have several problems to directly use them. First, there are many playlist titles that cannot be considered as a phrase. Melon includes 27,420 playlists with empty titles which is 18.4\% of the total playlists. In the case of MPD, 646,868 playlists have titles with a single word which amount to 64.7\% of the total. In addition, there are playlist titles that have multiple tokens but not a phrase, for example, ``G e o r g e W i n s t o n e'' and ``beyonce - 4''. Finally, some playlists have zero or few songs which are typically not considered as a playlist. The statistic of the two datasets is summarized in the \textit{Original} column of Table \ref{data_stat}.


We reorganized the two datasets with the same criteria to improve the quality of data samples for playlist title generation. First, we gather all playlists provided by each dataset. In the case of Melon, we merged the provided train, validation, and test set into one, and then filtered out some playlists with three criteria. First, the number of title tokens should be more than 3. Second, the number of tracks should be more than 10. Third, the average character length of title tokens should be more than 3. 

Finally, the filtered dataset is split by the number of title tokens. Playlists with the same number of title tokens are randomly split with a ratio of 8:1:1 and merged among different numbers of tokens subsequently to form train, validation, and test set. As a result, the statistics of data was changed as as shown in \textit{Filtered} column of Table \ref{data_stat}. The longer average character length and title length indicate the portion of playlist titles in phrases within the dataset has increased.


\begin{table}[!t]
\resizebox{\linewidth}{!}{%
\begin{tabular}{|l|l|c|c|}
\hline
\textbf{Dataset} & \textbf{Statistic} & \textbf{Original} & \textbf{Filtered} \\ \hline
\multirow{7}{*}{\begin{tabular}[c]{@{}l@{}}Melon\\ Playlist\end{tabular}} & Playlist Number & 148,826 & 51,723 \\ \cline{2-4} 
 & Unique Track Number & 649,092 & 430,746 \\ \cline{2-4} 
 & Unique Title Number & 115,318 & 50,296 \\ \cline{2-4} 
 & Unique Word Number & 88,524 & 56,296 \\ \cline{2-4} 
 & Average Char Length & 2.8 & 3.6 \\ \cline{2-4} 
 & Average Title Length & 3.6 & 4.7 \\ \cline{2-4} 
 & Average Track Length & 39.7 & 46.2 \\ \hline
\multirow{7}{*}{\begin{tabular}[c]{@{}l@{}}Spotify\\ Million\\ Playlist\end{tabular}} & Playlist Number & 1,000,000 & 50083 \\ \cline{2-4} 
 & Unique Track Number & 2,262,292 & 402,523 \\ \cline{2-4} 
 & Unique Title Number & 17,381 & 1,859 \\ \cline{2-4} 
 & Unique Word Number & 11,146 & 1,886 \\ \cline{2-4} 
 & Average Char Length & 5.2 & 4.2 \\ \cline{2-4} 
 & Average Title Length & 1.4 & 3.4 \\ \cline{2-4} 
 & Average Track Length & 66.3 & 66.3 \\ \hline
\end{tabular}
}
\vspace{-1mm}
\caption{Compare statistic of datasets. After filtering, as the average title length increases, it can be seen that the noise of each phrase has been removed.}
\label{data_stat}
\end{table}

\section{Playlist Title Generation}

\subsection{Encoder-Decoder Model}
The model for playlist title generation is composed of an encoder and a decoder. The goal of the model is to find a title word sequence $y$ that maximizes the conditional probability of $y$ given a source track ID sequence $x$. The encoder reads a track ID sequence $x = (x_{1},..,x_{n})$, represents track ID as an embedding vector using random initialized embedding matrix $ E \in \mathbb{R}^{|V| \times d}$, and transforms it to hidden states $z = (z_{1}..,z_{n})$. The decoder takes these hidden states as a context input and outputs a summary $y = (y_{1}..,y_{m})$. At each step the model is auto-regressive, consuming the previously generated symbols as an additional input when generating the next. During training, we used the softmax cross-entropy loss. The encoders and decoders can be RNN \cite{bahdanau2014neural}, Convolutional Neural Network (CNN) \cite{gehring2017convolutional} or self-attention layer \cite{vaswani2017attention}. In this paper, we compare the RNN model and the Transformer model composed of self-attention layers. 

\vspace{3mm} \noindent \textbf{RNN Model:}
Our baseline model corresponds to the neural machine translation model used in \cite{bahdanau2014neural}. The encoder consists of bidirectional Gated Recurrent Unit (GRU) \cite{chung2014empirical}, while the decoder consists of a uni-directional GRU with the same hidden-state size as that of the encoder, and an attention mechanism over the source-hidden states and a soft-max layer over the target vocabulary to generate words. 

\vspace{3mm}  \noindent \textbf{Transformer:}
The encoder and decoder are composed of multi-head self-attention layers and position-wise fully connected feed-forward network with a residual connection and a layer normalization.\cite{vaswani2017attention}. The transformer views the encoded representation of the input as a set of key-value pairs and both the keys and values are the encoder hidden states. In the decoder, the previous output is compressed into a query and the next output is produced by mapping this query and the set of keys and values. The output of self-attention layer is a weighted sum of the values, where the weight is calculate by the dot-product the query with all the keys.

\subsection{Ignoring the Order in Track Sequences}
One of the characteristics of playlists is that the order of tracks in a playlist is generally not important. This feature can be exploited for data augmentation. In this paper, we propose two different techniques. The first is sequence shuffling which randomly changes the order of tracks in the same playlist. 
The second is to remove the positional encoding of the encoder. According to the loss of position information, the model can recognize the track sequence except for the sequence information of the data. On the other hand, the decoder applies the positional encoding to the word sequence for title generation. We applied the two techniques independently, because, when the positional encoding is removed, the transformer model does not recognize the input sequence differently regardless of shuffling.

\begin{table}[!t]
\resizebox{\linewidth}{!}{
\begin{tabular}{l|cc|cc}
\midrule
\multirow{2}{*}{Model} & \multicolumn{2}{c|}{Melon} & \multicolumn{2}{c}{MPD} \\ 
 & Val NLL & Test NLL & Val NLL & Test NLL \\ \midrule
RNN Model & 7.482 & 7.384 & 2.453 & 2.357 \\ \midrule
Transformer & 7.150 & 7.124 & 1.821 & 1.805 \\ 
+ shuffle aug & \textbf{6.952} & \textbf{7.019} & \textbf{1.543} & \textbf{1.502} \\ 
+ delete pos & 7.036 & 7.099 & 1.552 & 1.538 \\ \midrule
\end{tabular}
}
\vspace{-3mm}
\caption{Validation and test NLL for melon and spotify million playlist dataset. The \textbf{shffle aug} means data augmentation through shuffling the input track sequence, and \textbf{delete pos} means that delete encoder's positional encoding in vanilla transformer.}
\label{nll}
\end{table}

\begin{table*}[]
\small
\centering
\resizebox{\textwidth}{!}{%
\begin{tabular}{l|l|l}
\midrule
Playlist ID & 49169 & 22501 \\ \midrule
Input Tracks & \begin{tabular}[c]{@{}l@{}}Swear by Inc., Millionairess by Inc., \\ Her Favorite Song (w/Crossfade) by Mayer Hawthorne, \\ Dontcha by The Internet, All I Do by Majid Jordan, \\ Her by Majid Jordan, Us by MOVEMENT, \\ The Place by Inc., Coffee (Feat. Wale) by Miguel, \\ Under Control by The Internet, \\ Somthing`s Missing by The Internet, \\ Ocean Drive by Duke Dumont, Drive by Dornik, \\ Make It Work by Majid Jordan, \\ Jump Hi (Feat. Childish Gambino) by Lion Babe, \\ Treat Me Like Fire by Lion Babe, sHe by ZAYN,\\ Hallucinations by dvsn, \\ Dapper (Feat. Anderson .Paak) by Domo Genesis, \\ Bone + Tissue by Gallant, Miyazaki by Gallant, \\ ...\end{tabular} & \begin{tabular}[c]{@{}l@{}}Take Five by Michel Camilo, \\ Angelina by Tommy Emmanuel, \\ Monk`s Dream (Live) by Martin Reiter, \\ Stairway To Love by George Benson, \\ Birdsong by Tommy Flanagan, \\ Come Fly With Me by Frank Sinatra, \\ Gemini by Chick Corea, Cheesecake by Dexter Gordon, \\ Kathy by Horace Silver, Love Me by The Little Willies, \\ Perdido by Earl Hines, `Round Midnight by Hank Jones, \\ I Just Called To Say I Love You by Harry Allen,\\  Killing Me Softly With His Song by Harry Allen, \\ Let`s Fall In Love by Diana Krall, \\ Flight To Jordan by Duke Jordan, \\ Quizas Quizas Quizas by Lisa Ono, \\ ...\end{tabular} \\ \midrule
\begin{tabular}[c]{@{}l@{}}Ground Truth\end{tabular} & late night drive & \begin{tabular}[c]{@{}l@{}}가을밤 로맨틱 재즈곡들\\ romantic jazz songs for an autumn night \color{gray}{(translated)}\end{tabular} \\ \midrule
RNN Model & {\begin{tabular}[c]{@{}l@{}}몽환적인 r\&b r\&b \\ dreamy r\&b r\&b \color{gray}{(translated)}\end{tabular}} & jazz jazz jazz jazz \\ \midrule
Transfomer & \begin{tabular}[c]{@{}l@{}}생생한 고음질로 만나는 hi-fi 위클리 12월 16일 vol 1\\ lively and high-quality sound in hi-fi weekly December 16th \\ vol 1 \color{gray}{(translated)}\end{tabular} & \begin{tabular}[c]{@{}l@{}}카페에서 듣는 음악들\\ music in a cafe \color{gray}{(translated)}\end{tabular} \\ \midrule
\begin{tabular}[c]{@{}l@{}}Transfomer\\ + shuffle aug\end{tabular} & \begin{tabular}[c]{@{}l@{}}들을수록 좋은 세련되고 감각적인 pop\\ stylish and sensual pop that you feel better as you listen more \\  \color{gray}{(translated)}\end{tabular} & \begin{tabular}[c]{@{}l@{}}카페에서 듣는 감각적인 재즈\\ sensational jazz in a cafe \color{gray}{(translated)}\end{tabular} \\ \midrule
\begin{tabular}[c]{@{}l@{}}Transfomer\\ + delete pos\end{tabular} & \begin{tabular}[c]{@{}l@{}}내가 좋아하는 노래\\ my favorite song \color{gray}{(translated)}\end{tabular} & \begin{tabular}[c]{@{}l@{}}카페에서 듣는 잔잔한 음악\\ calm music in a cafe \color{gray}{(translated)}\end{tabular} \\ \midrule
\end{tabular}%
}
\caption{Inference example from the melon playlist test dataset. Reference refers to the ground turth of the dataset. Each first line is a generation result, and the second line is a phrase translated from Korean to English. Source means input track sequence, and track index over 15 are excluded.}
\label{quanti}
\end{table*}

\subsection{Training Details}
We fixed the number of layers of encoder and decoder to two and 128 embedding dimensions and 256 hidden dimensions for fair comparison in the two types of encoder-decoder models. We trained the model using a single GPU. We optimized the model using the Adam optimizer \cite{kingma2014adam} with a 0.005 learning rate, and 0.0001 learning rate decay for all models and datasets. We used a batch size of 64 and randomly shuffled the training data at every epoch. We used early stopping on the validation set, monitoring with the validation loss, and used the best model on the validation set to report all performance numbers. 

\section{Results and Discussion}
\subsection{Quantitative Results}
We used negative log-likelihood (NLL) as an evaluation metric for the models. Table \ref{nll} lists the NLL values for the RNN and Transformer models on the two datasets.  
The result shows that the Transformer models generally outperform the baseline RNN model on both datasets. In addition, the Transformer models that ignore the order of track sequence improve the performance further. Between the two techniques, shuffling augmentation has a slightly lower NLL value than deleting the positional encoding on on both datasets. This indicates the data augmentation approach that involves ignoring the order is more effective than simply removing the order information. 



\subsection{Qualitative Results}
Table \ref{quanti} shows two examples of title generation given an input track sequence. We can first see that the RNN models generate a short title. They even have repetitions of the genre words (e.g., r\&b, jazz). On the other hands, the Transformer models generates a natural phrase composed of more than 3 different words. An interesting result in the example on the left side is that the basic Transformer model has a very specific title which seems to be copied from data with strong context (``hi-fi weekly December 16th, vol. 1''). This problem is alleviated in the Transformer models with shuffle augmentation or deleting the position encoding.   


\section{Conclusions}
In this work, we propose music playlist title generation with a machine translation approach. There are several future directions to extend this work. First, we can try various track embedding vectors for the input sequence. For example, we can use tag prediction vectors or audio embedding vectors from music auto-tagging models or track embedding vector from matrix factorization of user listening data. Second, we should investigate more quantitative metrics to evaluate the models. The BLEU score used in machine translation may be a possibility in terms of accuracy but we should also consider the diversity of the generated playlist titles to provide rich expressions for music listeners. Finally, we need to have a user study designed systemically that compare different models of playlist title generation.     




\bibliography{nlp4MusA}
\bibliographystyle{nlp4MusA_natbib}

\end{document}